\documentclass[a4paper,oneside,twocolumn]{article}
\usepackage{cite}
\usepackage{hyperref}
\usepackage{amsmath,amssymb,amsfonts}
\usepackage[linesnumbered, ruled]{algorithm2e}
\usepackage[noend]{algpseudocode}
\usepackage{framed,multirow}
\usepackage{graphicx}
\usepackage{textcomp}
\usepackage{xcolor}
\usepackage[margin=2cm]{geometry}
\usepackage{float}

\begin{document}

\title{A Pseudo-labelling Auto-Encoder for unsupervised image classification 
}

\author{Aymene Mohammed Bouayed\footnote{Corresponding author --- email: \href{mailto:bouayedaymene@gmail.com}{\texttt{bouayedaymene@gmail.com}}} , Karim Atif \\
\textit{Computer science department} \\
\textit{University of Sciences and Technology Houari Boumediene }\\
Algiers, Algeria
\and
Rachid Deriche\\
\textit{Athena Project-Team, INRIA Sophia-Antipolis-M\'edit\'erran\'ee} \\
Sophia Antipolis, France
\and
Abdelhakim Saim\\
\textit{Institut de Recherche en Energie} 
\textit{Electrique de Nantes Atlantiques} \\
Saint-Nazaire, France 
}
\date{}

\maketitle

\begin{abstract}
In this paper, we introduce a unique variant of the denoising Auto-Encoder and combine it with the perceptual loss to classify images in an unsupervised manner. The proposed method, called Pseudo Labelling, consists of first applying a randomly sampled set of data augmentation transformations to each training image. As a result, each initial image can be considered as a pseudo-label to its corresponding augmented ones. Then, an Auto-Encoder is used to learn the mapping between each set of the augmented images and its corresponding pseudo-label. Furthermore, the perceptual loss is employed to take into consideration the existing dependencies between the pixels in the same neighbourhood of an image. This combination encourages the encoder to output richer encodings that are highly informative of the input's class. Consequently, the Auto-Encoder's performance on unsupervised image classification is improved in terms of stability, accuracy and consistency across all tested datasets. Previous state-of-the-art accuracy on the MNIST, CIFAR-10 and SVHN datasets is improved by 0.3\%, 3.11\% and 9.21\% respectively.
\end{abstract}

\vspace{0.4cm}

\textbf{\textit{Keywords} ---}
Auto-Encoder, Denoising Auto-Encoder, Perceptual Loss, Data Augmentation, Unsupervised Learning, Image Classification.

\section{Introduction}
Classification is one of the most important tasks in deep learning. It consists of identifying a trait in the input and assigning a label to it. The input could be an image, a video, a simple vector of values or else. 

Classification has many useful and valuable applications such as spam detection \cite{spam}, disease identification \cite{maladie}, particle discovery \cite{higgs} etc. Current deep learning technics are able to achieve outstanding performance on this task using supervised learning. However, the efficacy of these methods depends on the presence of labeled data which is very scarce. For this aim, the development of unsupervised and semi-supervised methods has seen an increasing interest. 

In order to benefit from the sheer amount of available unlabelled data, a lot of work has been done to improve the performance of deep learning models in the unsupervised learning context. Among these methods, in \cite{unsupervised-gan} it is proposed to concatenate the output of some of the convolutional layers of the GAN's discriminator (Generative Adverserial Network) and pass them through a linear classifier to infer the class of the input images. In the work of \cite{scae}, a Stacked Capsule Auto-Encoder has been used to break down an image into multiple objects. Then, according to the present objects in a scene, the class of the image can be deduced.

\begin{figure*}[!t]
\centering
\includegraphics[width=\textwidth]{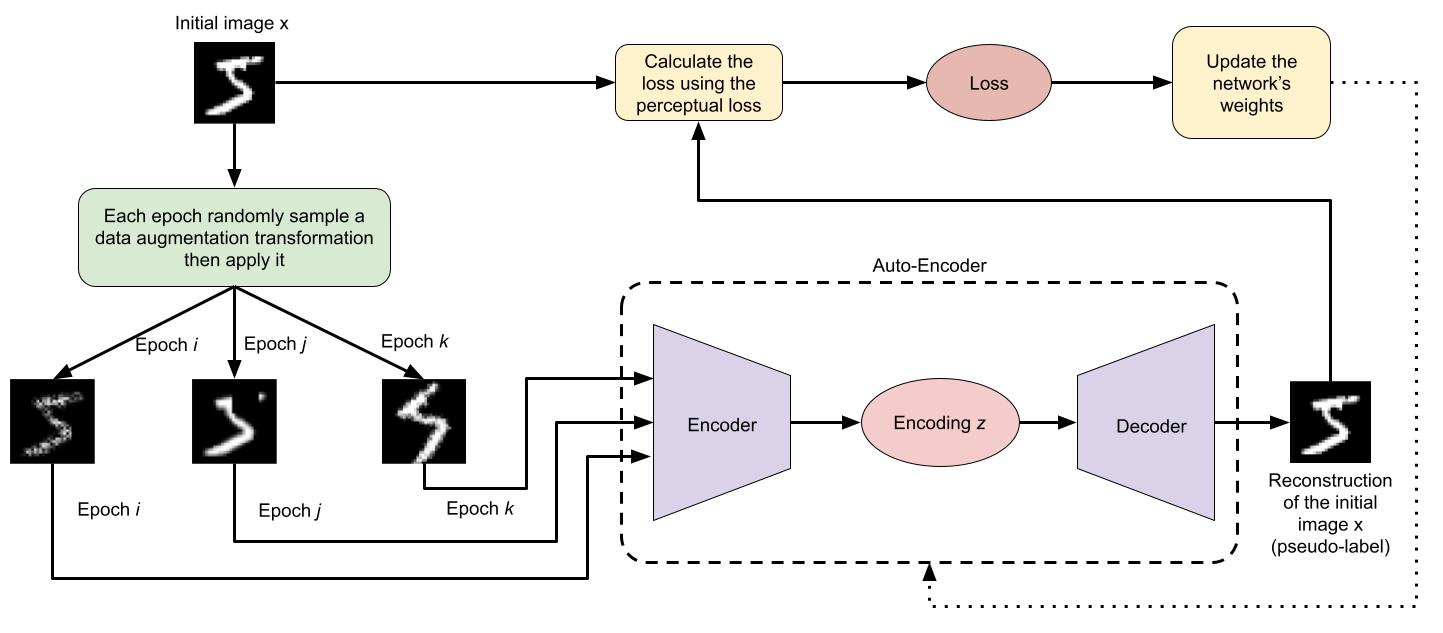}
\caption{Pseudo-Labelling Auto-Encoder method proposed in this work.}
\label{proposed system}
\end{figure*}

Auto-Encoders are neural networks that are trained to attempt to reconstruct the input that is presented to them and construct an internal representation of it \cite{deep-learning-book}. They have been used to different ends among them modelling \cite{modelling}, de-noising \cite{denoising} and more. 

The work of \cite{projet-guessoum} explored the use of Auto-Encoders to do unsupervised image classification and optimised the architecture of the network using the Genetic Algorithm. This work obtained satisfactory results on the MNIST dataset where 96.32\% accuracy has been achieved with an optimised Auto-Encoder that has fully connected layers.

Conventionally, Auto-Encoders use a pixel-wise loss function to calculate the difference between the input and the output. Thus, the pixel-wise loss does not take into consideration the dependencies between the neighbouring pixels of an image. To solve this problem, the perceptual loss was introduced by A. B. L. Larsen \& al. in \cite{first-perceptual-loss}. Its importance and impact have been evaluated for image reconstruction and generation where it demonstrated great potential. The perceptual loss uses a pre-trained model to capture the dependencies between the pixels of an image then applies the loss function to the output of this model.

The work of \cite{p-ae} explored the effect of the perceptual loss on the embeddings produced by the encoder in the task of object positioning and image classification. It has been found that the perceptual loss enriches the encodings produced w.r.t the semantic class of the input. Big improvements in terms of classification accuracy have been noticed over not using the perceptual loss. 

Similarly, the use of data augmentation has been shown as a powerful technique to improve the generalisation capabilities on neural networks in a supervised learning context \cite{survey-data-augmentation}.

Data augmentation is a technique where alterations are introduced into the training data in order to improve its diversity and allow the network to see the data from different perspectives during the training phase. In the recent work of T. Chen \& al. \cite{simclr} data augmentation was one of the most important ideas that allowed obtaining state-of-the-art performance on multiple datasets, surpassing the accuracy of neural networks having done the learning phase in a supervised manner. In the work of \cite{simclr}, data augmentation is applied to an image twice and then encoded using a large pre-trained model. The weights of the network are then refined so that the encodings of the image are the same regardless of the applied data augmentations.

Despite the effectiveness of data augmentation in the supervised and the semi-supervised contexts, this technique's effect on the embeddings produced by an Auto-Encoder in the unsupervised image classification task has not been explored yet, notably when used in conjunction with the perceptual loss. In this work, we aim to explore this combination.

To this end, we found that incorporating data augmentation in a simple way where the Auto-Encoder reconstructs its input, which is an image with data augmentation, does not improve the embeddings produced. As a result, we introduce an effective method for the integration of data augmentation called the \textit{Pseudo-Labelling} method. This method is a variant of a denoising Auto-Encoder and consists of creating a correspondance between data augmented images and the original image (see Figure \ref{proposed system}). Combining the proposed method with an Auto-Encoder trained with the perceptual loss results in improvements in both the stability and the accuracy of the unsupervised classification. Consequently, the proposed method incorporates data augmentation in a way that forces the encoder to better understand the input, look past the alterations applied and create more concise encodings.

Subsequently, we summarise our contributions in this work as follow:

\begin{itemize}
\item We emphasise the improvements that the perceptual loss brings when compared to a simple Auto-encoder w.r.t the unsupervised images classification accuracy. 
\item We propose a novel combination of a variant of a denoising Auto-Encoder and the perceptual loss called \textit{Pseudo-Labelling} that leads the encoder to improve the quality of the outputted embeddings w.r.t unsupervised image classification.
\item We compare positively the performance of the Pseudo-Labelling method to 8 different unsupervised learning methods including Auto-Encoder and none Auto-Encoder based methods \cite{unsupervised-gan} \cite{scae}  \cite{p-ae} \cite{adc} \cite{ae} \cite{imsat}  \cite{iic}.
\end{itemize}

The outline of this paper is as follows. In Section 2, we present the basic concepts of the Auto-Encoders then we detail the main idea of Auto-Encoders trained using the perceptual loss. In Section 3, we deal with the data augmentation method applied to images. Section 4 introduces two different ways of incorporating data augmentation in an Auto-Encoder that is trained using the perceptual loss. The first one uses data augmentation in a simple way, whereas the second one (the Pseudo-Labelling method) harnesses data augmentation in a beneficial way that urges the network to learn better encodings. In Section 5, we present the results of our experiments. Section 6 compares the work presented in this paper with a variety of Auto-Encoder and none Auto-Encoder based methods in the task of unsupervised image classification. Our conclusions and perspectives for future work are presented in Section 7.

\section{Auto-Encoder}
\subsection{Basic Concepts}
An Auto-Encoder (B-AE: \textit{Basic Auto-Encoder}) is a special kind of neural network architecture that is composed of two neural networks an encoder and a decoder. The encoder and the decoder are stacked one on top of the other and trained at the same time. The goal of the encoder is to produce a concise encoding $z$ that keeps all the important information about the input $x$. The decoder's task is to use the encoding $z$ and reconstruct the input the best way possible \cite{deep-learning-book}.

\begin{equation}
\begin{aligned}
   z&=encoder(x), \\
x^\prime&=decoder(z).
  \end{aligned}
\end{equation}

\begin{figure}[H]
\centering
\includegraphics[width=0.49\textwidth]{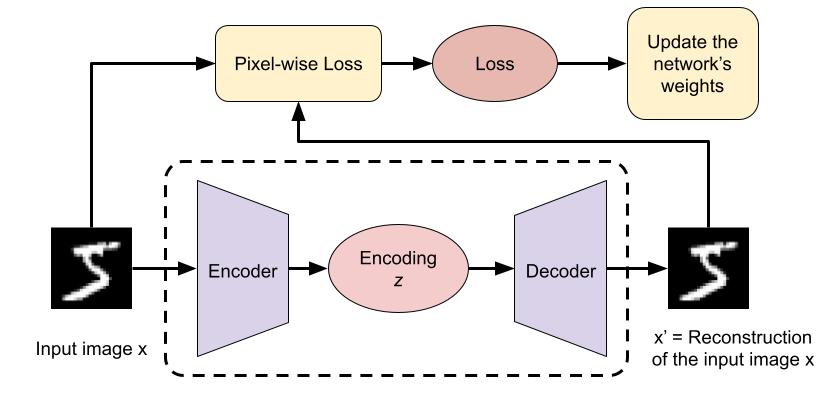}
\caption{A chart of a Basic Auto-Encoder where the input image $x$ is passed through the encoder to obtain an encoding $z$. Then, the encoding $z$ is passed through the decoder to generate $x^\prime$ which is a reconstruction of $x$. Finally, $x$ and $x^\prime$ are compared using a pixel-wise loss function.}
\label{base auto encoder}
\end{figure}

The whole network is trained using the back-propagation algorithm. While the loss is computed by comparing the input image $x$ to its reconstruction $x^\prime$ using a loss function $\ell$ such as the MSE (Mean Squared Error) loss (See Figure \ref{base auto encoder} and Algorithm \ref{b-ae algo}).

\begin{algorithm}[ht]
 \caption{B-AE main learning algorithm.}
 \label{b-ae algo}
\SetAlgoLined
\textbf{input: } batch size $N$, number of epochs $epochs$

epoch $\leftarrow$ 1

\While{epoch $\leq$ $epochs$}{

\For{\textbf{each} sampled minibatch $\{x_i\}_{k=1}^N$ }{

\For{\textbf{all} $k \in \{1 ,\dots, N \}$}{

$z_k$ $\leftarrow$ $encoder(x_k)$

$x^\prime_k$ $\leftarrow$ $decoder(z_k)$

}

$\mathcal{L} = \sum_{k=1}^N \ell(\boldsymbol{x^\prime_k},\boldsymbol{x_k})$

Update the $encoder$'s and the $decoder$'s weights to minimise $\mathcal{L}$
}

epoch $\leftarrow$ epoch + 1

}
\textbf{return} $encoder$ and $decoder$
 \end{algorithm}

\subsection{Auto-Encoder with Perceptual Loss}
Convolutional layers first introduced in the work of \cite{cnn}, rely on the fact that pixels in an image are not independent entities but they depend on the neighbouring pixels. This idea allows the extraction of more information from the images by capturing these dependencies and harnessing them to perform better on the task at hand.

\begin{figure}[H]
\centering
\includegraphics[width=0.49\textwidth]{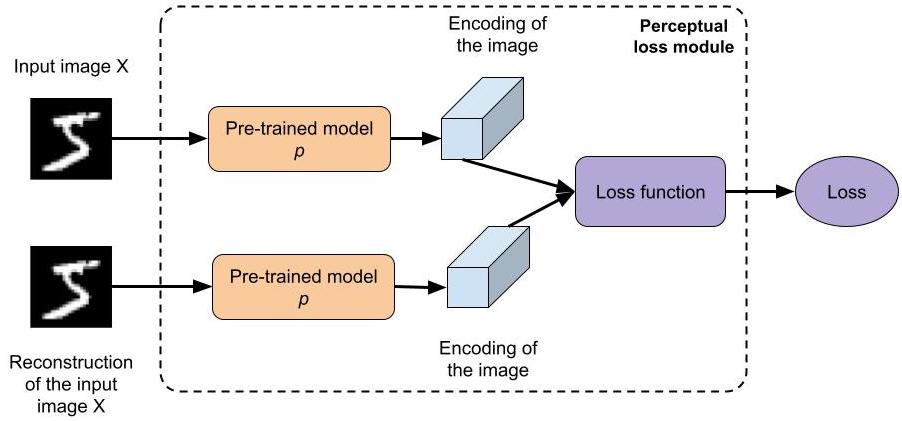}
\caption{Content of the perceptual loss module. In an Auto-Encoder that uses the perceptual loss, the pixel-wise loss module of Figure \ref{base auto encoder} is replaced by this module where first the original image $x$ and its reconstruction $x^\prime$ are passed through a pre-trained model $p$. This yields encodings which contain the important information in the images and the extracted dependencies between the pixels. Then, a loss function such as the MSE Loss is used to compare the encodings pixel-wise.}
\label{pipeline perceptual loss}
\end{figure}

This concept is also used in the perceptual loss of Auto-Encoders \cite{first-perceptual-loss} \cite{p-ae}. Where in this context too the value of a pixel depends on the value of the neighbouring pixels. So, the loss function $\ell$ should not be applied pixel-wise between the input $x$ and its reconstruction $x^\prime$ as done in the B-AE:
\begin{equation}
\label{loss-bae}
\mathcal{L} = \sum \ell(x^\prime,x).
\end{equation}
Instead, $x$ and $x^\prime$ are passed through a portion of a pre-trained model $p$ that captures the relationships between the pixels in the same neighbourhood. Then, the loss function $\ell$ is applied to the output of the pre-trained network $p$. As a result, for the Auto-Encoder with perceptual loss (P-AE), equation (\ref{loss-bae}) and the corresponding formula in line 9 of the Algorithm \ref{b-ae algo} should be replaced with the following formula:
\begin{equation}
\mathcal{L} = \sum \ell(p(x^\prime),p(x)).
\end{equation}
This allows to compare regions of the input image $x$ with their correspondent regions of the reconstructed image $x^\prime$. Figure \ref{pipeline perceptual loss} illustrates the perceptual loss module.

The impact of the P-AE on the quality of the embeddings produced by the encoder has been discussed in the work of \cite{p-ae}. It has been found that the perceptual loss greatly enriches the produced latent vector $z$ allowing a higher accuracy in unsupervised classification.

\section{Data Augmentation}
Data augmentation is a technique that is heavily used in the field of computer vision. It allows the neural network to build an internal representation of the data that is more robust to small changes in the input and better generalise to new data thus obtaining a higher accuracy \cite{survey-data-augmentation}.

\begin{figure}[H]
\centering
\includegraphics[width=0.49\textwidth]{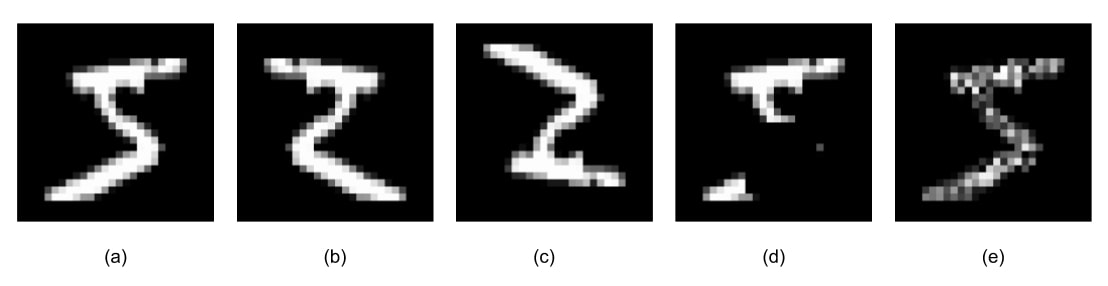}
\caption{Examples of data augmentation transformations on an image from the MNIST dataset. (a) represents the original image, (b) and (c) represent horizontally and vertically flipping the original image, (d) represents the cutout transformation and (e) represents the original image with added gaussian noise.}
\label{examples data augmentation}
\end{figure}

Data augmentation works through the introduction of new images into the training set by transforming the existing ones. The transformations that could be applied include but not limited to horizontally or vertically flipping an image, hiding parts of the image (cutout), adding gaussian noise to the image and so on (see Figure \ref{examples data augmentation}).

\section{Proposed Approach}
\subsection{Simple Data Augmentation Auto-Encoder}
In order to further improve the quality of the encodings produced by the P-AE, data augmentation could be incorporated into the Auto-Encoder. 

A Simple Data Augmentation Auto-Encoder (SDA-AE) can be used where a data augmentation transformation $t$ is applied to the input image $x$ to obtain a modified image denoted $f_S$
\begin{equation}
f_S = t(x).
\end{equation}

Then, the modified image $f_S$ is used as input and target of the P-AE (see algorithm \ref{SDA-AE algo}).

\begin{algorithm}[ht]
 \caption{SDA-AE main learning algorithm.}
 \label{SDA-AE algo}
\SetAlgoLined

\textbf{input: } batch size $N$, number of epochs $epochs$, a portion of a pre-trained model $p$

epoch $\leftarrow$ 1

\While{epoch $\leq$ $epochs$}{

\For{ \textbf{each} sampled minibatch $\{x_i\}_{k=1}^N$}{

\For{\textbf{all} $k \in \{1 ,\dots, N \}$}{

$t$ $\leftarrow$ a data augmentation transformation

$f_{S,k}$ $\leftarrow$ $t(x_k)$

$z_k$ $\leftarrow$ $encoder(f_{S,k})$

$x^\prime_k$ $\leftarrow$ $decoder(z_k)$

}

\textbf{\# Using the perceptual loss and applying it between the image with data augmentation and its reconstruction.}

$\mathcal{L} = \sum_{k=1}^N \ell(p(\boldsymbol{x^\prime_k}),p(\boldsymbol{f_{S,k}}))$

Update the $encoder$'s and the $decoder$'s weights to minimise $\mathcal{L}$.
}

epoch $\leftarrow$ epoch +  1

}

\textbf{return} $encoder$ and $decoder$
 \end{algorithm}
 
Even though the SDA-AE represents a viable solution to integrate data augmentation into the P-AE, it might be unstable during the training phase. This is due to the fact that there isn't a fixed target for each input. As a consequence, the quality of the important information stored in the encodings might fluctuate and not result in a better accuracy compared to the P-AE. For this aim we propose the PL-AE method in the next section.

\subsection{Pseudo Labelling Auto-Encoder}

The Pseudo Labelling Auto-Encoder (PL-AE) proposed in this work is an unsupervised method similar to denoising Auto-Encoders that imitates the mapping \textbf{input $\rightarrow$ label} used in the supervised learning context. 

\begin{figure}[ht]
\centering
\includegraphics[width=0.49\textwidth]{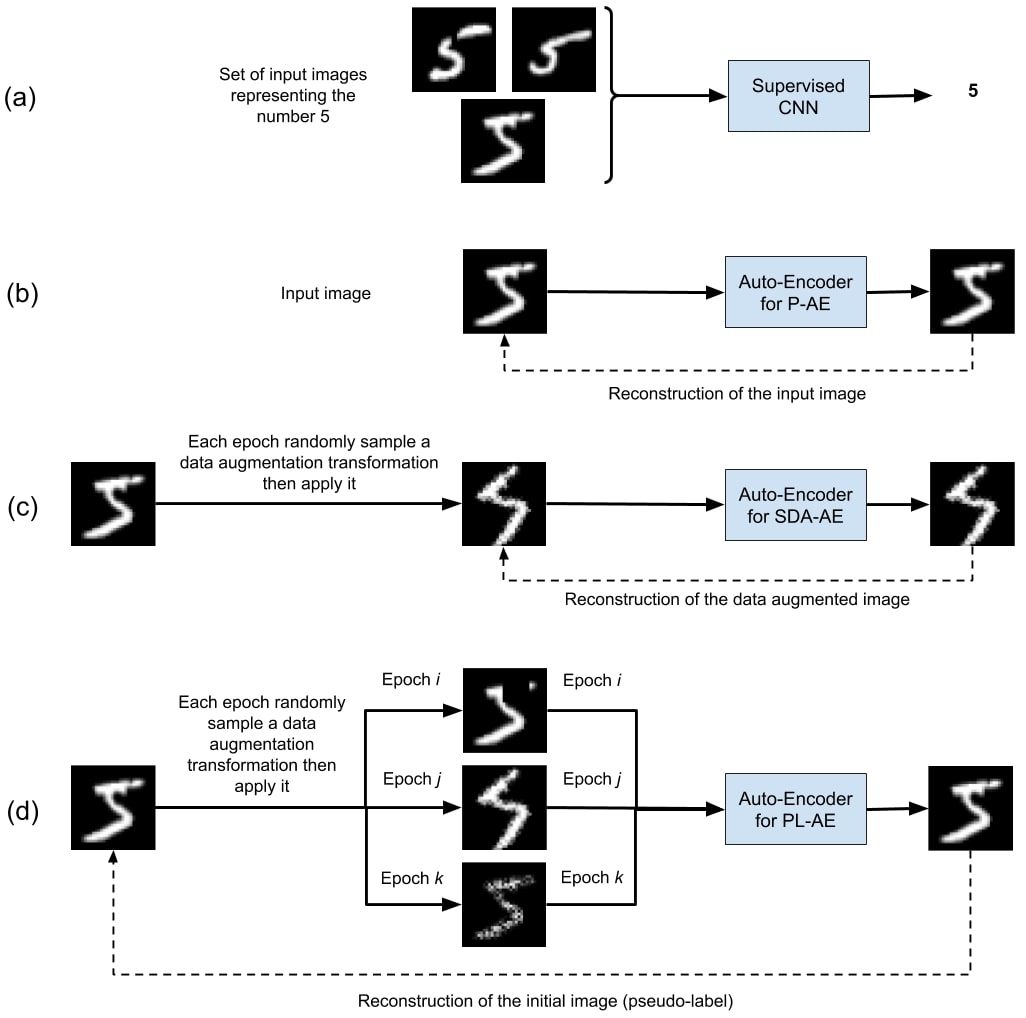}
\caption{PL-AE vs different Auto-Encoder approachs and a supervised CNN: The supervised CNN (a) maps the different images where 5 is drawn to the same label 5. The P-AE (b) and the SDA-AE (c) reconstruct their input images. Whereas the PL-AE (d) maps the different data augmented input images to the same initial image (pseudo-label).}
\label{pseudo labeling}
\end{figure}

To do so, various data augmentation transformations are applied to generate variations of the training images. This results in multiple data augmented images that correspond to each single initial training image. We can then consider the initial images as \textit{pseudo labels} to the sets of their corresponding data augmented images. After that we use an Auto-Encoder to learn the mapping \textbf{data augmented images $\rightarrow$ original image (pseudo-label)} (see Figure \ref{pseudo labeling}). Furthermore, the PL-AE makes use of the perceptual loss as it allows to construct richer encodings that provide accurate information w.r.t the image's class.

A key difference between the PL-AE and the denoising Auto-Encoder is that in the denoising Auto-Encoder usually one type data augmentation is applied, which is usually additive noise, and the task of the network is to circumvent it whereas in the PL-AE various data augmentation transformations are applied and the network tries to learn global features that characterise the input.
Also, PL-AE provides a stable target that is not affected by any data augmentation which allows for a much more stable learning as opposed to the SDA-AE.  

To apply the proposed method (PL-AE), firstly a data augmentation transformation $t$ is applied to the input image $x$ of the encoder. It should be noted that from epoch to epoch, the data augmentation transformation $t$ changes. This allows the creation of different variations of the original image and learn a better mapping between them and the original one
\begin{equation}
\begin{aligned}
f_{PL}&=t(x), \\
z&=encoder(f_{PL}).
\end{aligned}
\end{equation}

Then, the decoder takes the encoding $z$ and reconstructs the input $x$ (the one without any data augmentation) as opposed to the SDA-AE which reconstructs $f_S$
\begin{equation}
x^\prime=decoder(z).
\end{equation}

Where $x^\prime$ is the reconstruction of the original image $x$. 

Finally, the perceptual loss is used instead of the pixel-wise loss to compare $x$ with $x^\prime$ as it allows for a richer encoding $z$ as has been proven by \cite{p-ae}. 

As a result the training algorithm for the PL-AE is similar to the one of the SDA-AE with all instances of $f_{S,k}$ in Algorithm \ref{SDA-AE algo} replaced with $f_{PL,k}$, except for the loss function in line 12 which is replaced with the following formula:
\begin{equation}
\mathcal{L} = \sum_{k=1}^N \ell(p(x^\prime_k),p(\boldsymbol{x_k})).
\end{equation}
The complete PL-AE system is illustrated in Figure \ref{proposed system}.

\vspace{0.3cm}

It's important to note that the P-AE, SDA-AE and PL-AE are unsupervised methods since during the learning phase no labels are used and the pre-trained model employed is only used in the perceptual loss module. Also, in this work the pre-trained model used has been trained on the ImageNet dataset \cite{imagenet} which is different than the tested datasets. To have a fully unsupervised system, an alternative to obtain the pre-trained model can be to train a GAN say on the ImageNet dataset \cite{imagenet}. Then, the first layers of the descriminator of the GAN can used. The study of the best pre-trained model for the perceptual loss and the way to get it is outside the scope of this paper which focuses on proving the efficacy of the PL-AE method.

\section{Experiments and Results}
In this section we present our experiments to validate the PL-AE approach proposed in this work. 

\subsection{Datasets}

This work makes use of three different classification datasets which are MNIST \cite{cnn}, CIFAR-10 \cite{cifar10} and SVHN \cite{svhn}. 
\vspace{0.3cm}

\subsubsection{MNIST}
\begin{figure}[h]
\centering
\includegraphics[width=0.43\textwidth]{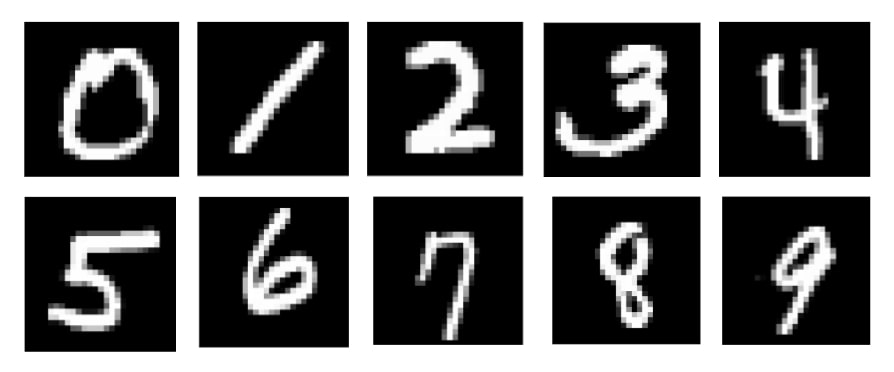}
\caption{Sample images from the MNIST datasets.}
\label{sample-mnist}
\end{figure}

The MNIST dataset is a collection of grey scale images of hand written digits. Each image has a size of $28\times28$ and only one channel. The dataset is divided into 60,000 images for the training set and 10,000 images for the test set. The goal in this dataset is to classify the images according to the digit that is drawn  (see Figure \ref{sample-mnist}) \cite{cnn}. 

In our work, before using the images of this dataset, we rescale them up to a size of $32\times32$, then we duplicate the image three times along the channels axis to have the same image but with three channels.

\subsubsection{CIFAR-10}
\begin{figure}[h]
\centering
\includegraphics[width=0.43\textwidth]{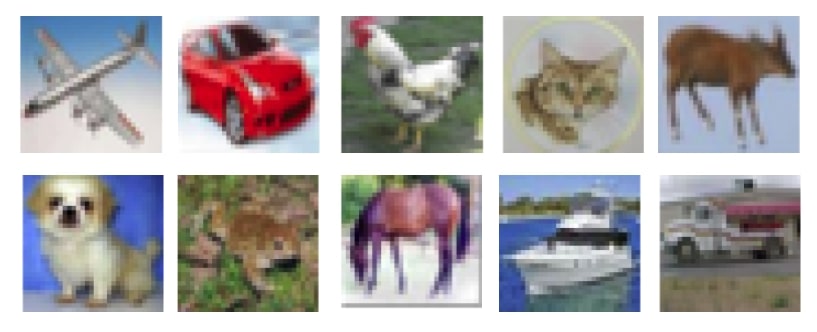}
\caption{Sample images from the CIFAR-10 datasets.}
\label{sample-cifar}
\end{figure}

The CIFAR-10 dataset is a collection of $32\times32$ coloured images of objects and animals. It englobes 50,000 images for the training set and 10,000 images for the test set. The goal in this dataset is to identify the object or the animal in the image (see Figure \ref{sample-cifar}) \cite{cifar10}.

\subsubsection{SVHN}
\begin{figure}[h]
\centering
\includegraphics[width=0.43\textwidth]{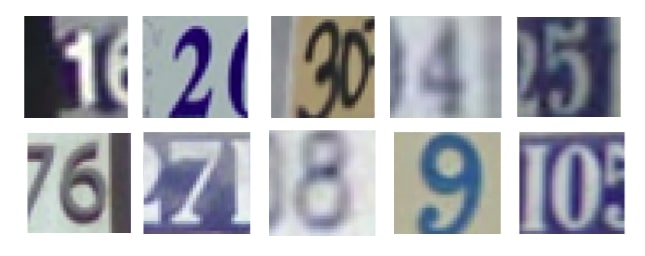}
\caption{Sample images from the SVHN datasets.}
\label{sample-svhn}
\end{figure}

The SVHN (Street View House Numbers) dataset is composed of 73,257 training images and 26,032 test images. Each image has a size of 32x32 and represents a digit from a house number, and the goal is to identify this digit (see Figure \ref{sample-svhn}) \cite{svhn}.

\vspace{0.3cm}

Finally, for each image of all the three datasets we duplicate it into a $2\times2$ grid to get an image of $64\times64$ pixels as it has been done by \cite{p-ae}. 

\subsection{Training Environment}
To implement the proposed method several libraries have been used namely:

\begin{itemize}
\item \textbf{The Pytorch library} \cite{pytorch} which provides implementations of different neural network layers, optimisers and pre-trained models. It has been used for the implementation of the perceptual loss and the different Auto-Encoders in this work.
\item \textbf{The Torchvision library} \cite{torchvision} which provides an easy access to several benchmark datasets and a multitude of data augmentation transformations. It has been used to get the datasets that have been tested in this work and apply data augmentation to them.
\item \textbf{The Sci-kit learn library} \cite{sklearn} which provides an implementation of linear classifiers \cite{svm} and the t-SNE algorithm \cite{tsne}. Where the latter has been used to cast the encodings from a high dimensional space to 2D to be plotted, and the linear classifier has been used to calculate the accuracy of the encodings generated by the Auto-Encoders for the different tested datasets.
\item \textbf{The Matplotlib library} \cite{matplotlib} which is a general purpose plotting library on Python. It has been used to draw the scatter plots of the obtained encodings.
\end{itemize}

Nvidia Tesla GPUs provided on the Google Colaboratory website which include T4, P4, P100 and K80 GPUs with 12Go of VRAM were used to train the different models. It takes for each epoch around 2, 3 or 4 minutes to train and evaluate the accuracy of an Auto-Encoder for the MNIST, CIFAR-10 and SVHN datasets respectively.


\subsection{Training Parameters}
\begin{figure}[ht]
\centering
\includegraphics[width=0.49\textwidth]{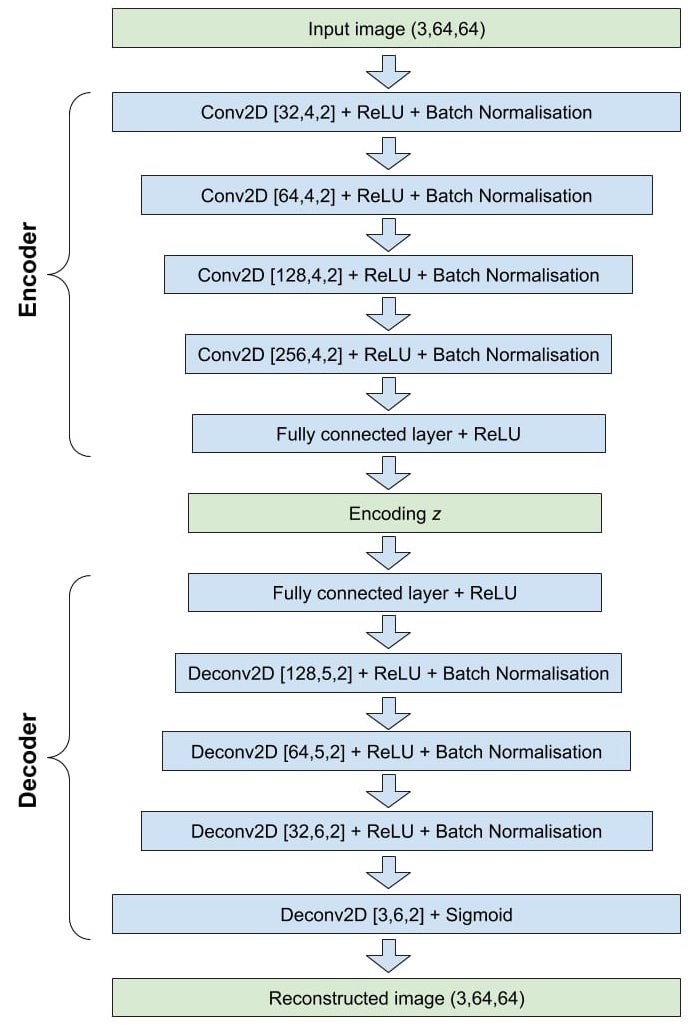}
\caption{Backbone architecture of the Auto-Encoder used in this work. Conv2D stands for the 2D convolution operation with [number of kernels, size of kernels, stride] parameters, ReLU for Rectified Linear Unit activation function, Deconv2D for the 2D deconvolution operation with [number of kernels, size of kernels, stride] parameters, and Sigmoid for the sigmoid activation function.}
\label{archi}
\end{figure}

All the Auto-Encoder training methods (B-AE, P-AE, SDA-AE and PL-AE) have been implemented using the same backbone architecture, similar to the one illustrated in Figure \ref{archi} and used by \cite{p-ae}.

Each Auto-Encoder has been trained for 90 epochs using the MSE loss function and Adam optimiser with a learning rate of lr=0.001, $\beta_1$=0.9 and $\beta_2$=0.999  \cite{adam} and a batch size of 100. 

\begin{figure}[ht]
\centering
\includegraphics[width=0.4\textwidth]{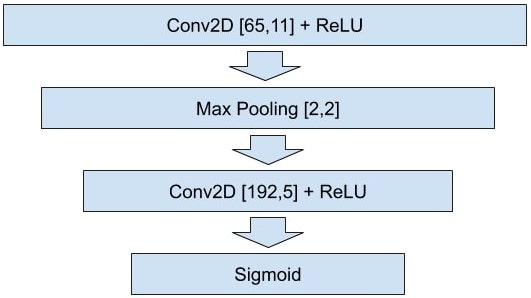}
\caption{The used portion of the Alexnet network in the perceptual loss module.}
\label{alexnet}
\end{figure}

For the models making use of the perceptual loss, the pre-trained model selected to test these methods, denoted as $p$ in Algorithm \ref{SDA-AE algo}, is the portion of the Alexnet model \cite{alexnet} pre-trained on the ImageNet dataset \cite{imagenet} showcased in Figure \ref{alexnet} which consists of the first three layers of this network. This portion is the same as the one used in \cite{p-ae} and in \cite{perceptual-2}.

Also, since the Alexnet model takes as input images of size 224$\times$224 and the images of the datasets used in this work are of size 64$\times$64, we pad the images with zeros all around until we get the desired size.

\subsection{Accuracy Evaluation}

The performance of the Auto-Encoders w.r.t the classification accuracy is computed by training a linear classifier on the training images encodings and their corresponding labels. Then, the linear classifier is evaluated on the test images encodings and their corresponding labels. Finally, the obtained accuracy is reported, as done in \cite{scae}. No data augmentation is used during this phase (see Algorithm \ref{accuracy evaluation}).

\begin{algorithm}[ht]
 \caption{Accuracy evaluation algorithm.}
 \label{accuracy evaluation}
\SetAlgoLined
\textbf{input: } training images and their labels ($train\_img$, $train\_labels$), test images and their labels ($test\_img$, $test\_labels$), The encoder part from the Auto-Encoder to be tested $encoder$, linear classifier $lin$.

train\_encoding $\leftarrow$ $encoder$($train\_img$)

test\_encoding $\leftarrow$ $encoder$($test\_img$)

$lin$.fit(train\_encoding, $train\_labels$)

accuracy $\leftarrow$ $lin$.score(test\_encoding, $test\_labels$)

\textbf{return} accuracy

 \end{algorithm}
 
\subsection{Data Augmentation Transformations Selection}
In order to find the data augmentation transformations that yield the best unsupervised classification accuracy, we studied different ones for each dataset using the proposed PL-AE method. 

We made sure to consider transformations that are semantically correct to apply then tested them. This is to ensure that the distribution of the obtained images is not far from the one of the train and test images. For example, we did not use vertical and horizontal flips for the datasets containing numbers (MNIST, SVHN) as their combination for the number 6 will give the number 9. Also, since denoising Auto-Encoders rely on the assumption that the input image contains sufficient information to recover the original image, we made sure not to change the distribution of the image too much by not using extreme values for the transformations. A study of the impact of different intensities of data augmentation transformations can make a subject of a future work. Table \ref{transformations} summarises the appropriate transformations that have been evaluated for each dataset. The parameters of each transformation can be found in appendix \ref{params DA}.

\begin{table}[H]
\resizebox{0.49\textwidth}{!}{
\centering
\begin{tabular}{ c  c  c  c}
\hline
\textbf{Transformation} &  \textbf{MNIST} & \textbf{CIFAR-10} & \textbf{SVHN} \\ \hline
\textbf{Random rotation} & \checkmark & \checkmark & \checkmark \\ 
\textbf{Affine transformation} & \checkmark & \checkmark & \checkmark \\ 
\textbf{Crop} & \checkmark & \checkmark & \checkmark \\ 
\textbf{Cutout} & \checkmark & \checkmark & \checkmark \\ 
\textbf{Gaussian noise} & \checkmark & \checkmark & \checkmark \\ 
\textbf{Colour jitter} & - & \checkmark & \checkmark \\ 
\textbf{Gray scale} & - & \checkmark & \checkmark \\ 
\textbf{Horizontal flip} & - & \checkmark & - \\ 
\textbf{Vertical flip} & - & \checkmark & - \\  \hline
\end{tabular}
}
\caption{Data augmentation transformations that have been tested for each dataset. }
\label{transformations}
\end{table}

\begin{figure}[H]
\centering
\includegraphics[width=0.4\textwidth]{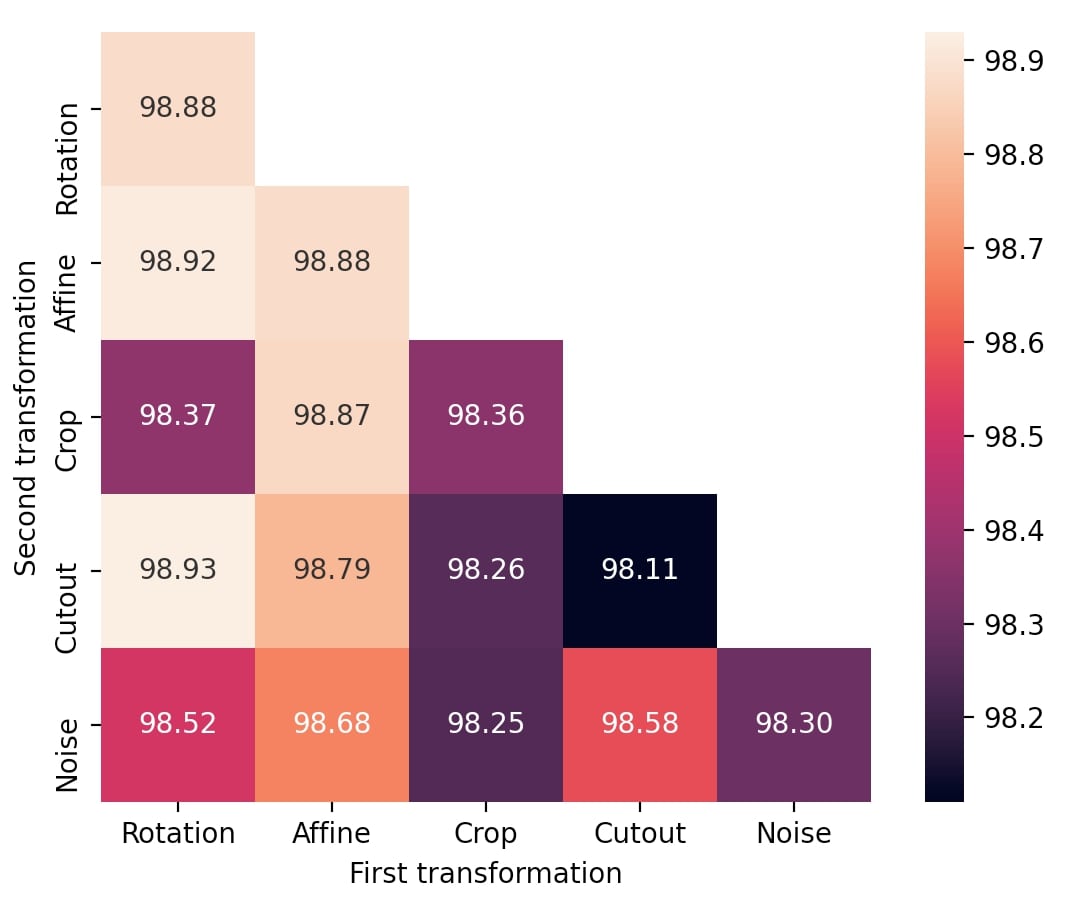}
\caption{Accuracies obtained for different data augmentations on the MNIST dataset.}
\label{heat map mnist}
\end{figure}

\begin{figure}[H]
\centering
\includegraphics[width=0.49\textwidth]{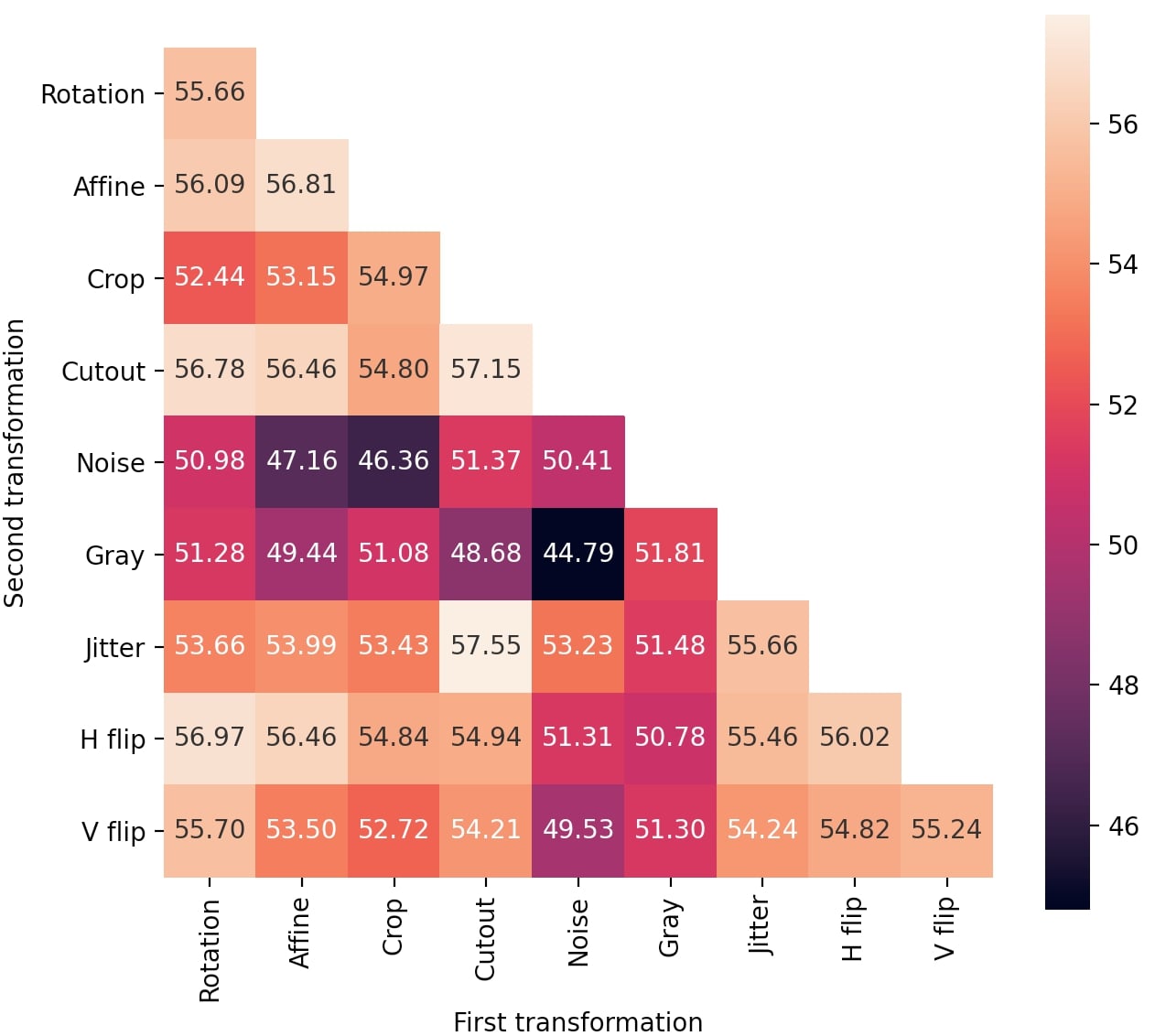}
\caption{Accuracies obtained for different data augmentations on the CIFAR-10 dataset.}
\label{heat map cifar10}
\end{figure}

\begin{figure}[H]
\centering
\includegraphics[width=0.45\textwidth]{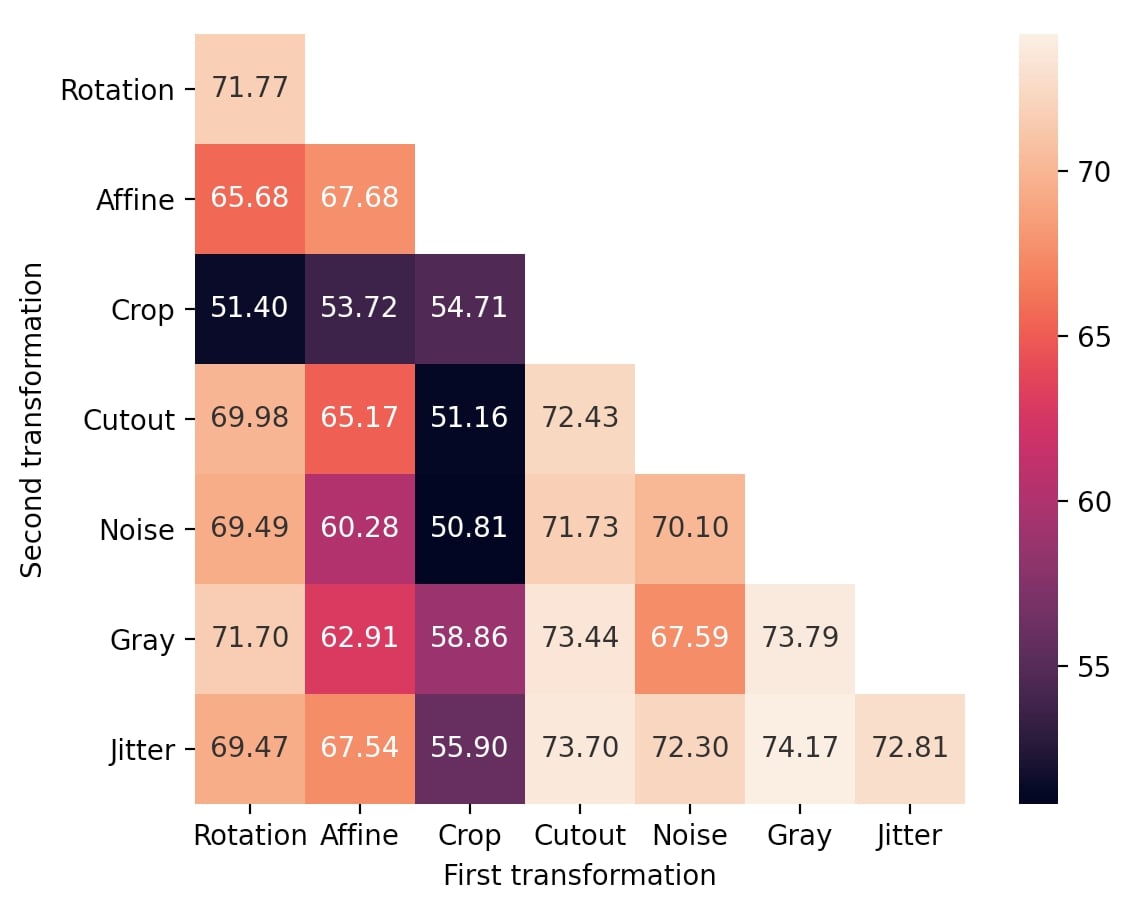}
\caption{Accuracies obtained for different data augmentations on the SVHN dataset.}
\label{heat map svhn}
\end{figure}

In addition, applying a combination of two different data augmentation transformations to the input images has been explored. This choice is motivated by the findings of T. Chen \& al. in \cite{simclr}. In \cite{simclr} it has been found that even though applying two different data augmentation transformations makes the learning process harder, it yields to richer encodings. The results obtained \footnote{The training parameters for the exploration of the different data augmentation transformations are the same as previously stated except for the number of epochs which was reduced to 30 epochs.} are reported in the heat maps of Figure \ref{heat map mnist}, \ref{heat map cifar10} and \ref{heat map svhn} which correspond to the MNIST, CIFAR-10 and SVHN datasets respectively. The diagonal represents when only one data augmentation is applied whereas the rest represents the use of two different data augmentations.

From figure \ref{heat map mnist} it can be seen that the accuracies obtained for the different data augmentations are close. This means that none of them has the potential to hinder the performance. As a result, sampling randomly from the list of the explored transformations for the MNIST dataset is possible and that's what we'll be doing in our following experiments on the MNIST dataset.

Figure \ref{heat map cifar10} and \ref{heat map svhn} show that several transformations did not yield high accuracies such as the gaussian noise for the CIFAR-10 dataset and the crop transformation for the SVHN dataset. So, if we sample randomly from a list of all possible transformations as done on the MNIST dataset there will be a performance degradation. To counter this problem and maximise the unsupervised classification accuracy, for our following experiments we sample from a list containing only the top-10 transformations for each dataset whenever data augmentation is used (See Appendix \ref{top-10}).

\subsection{Results}
\label{results}
After training different Auto-Encoders with different training methods (B-AE, P-AE, SDA-AE, PL-AE) we report on the column B-AE, P-AE, SDA-AE, PL-AE of the Tables \ref{results mnist}, \ref{results cifar} and \ref{results svhn} the best accuracies obtained. 

\begin{itemize}
\item B-AE column represents a basic Auto-Encoder which uses a pixel-wise loss and has as input and target the images $x$ without any data augmentation.
\item P-AE column \cite{p-ae} represents an Auto-Encoder which uses the perceptual loss and has as input and target the images $x$ without any data augmentation.
\item SDA-AE column represents an Auto-Encoder which uses the perceptual loss and has as input and target the images with data augmentation $F_S$.
\item PL-AE column represents an Auto-Encoder which uses the perceptual loss and the Pseudo-Labelling method.
\item CNN column represents the performance of a supervised CNN (Convolutional Neural Network) \footnote{The CNN was trained in a supervised manner using the cross entropy loss function. Its architecture is composed of the architecture of the used encoder (showcased in Figure \ref{archi}) combined with an MLP (Multi-Layer Perceptron) classification head.}. This column is used as reference to compare the unsupervised methods (B-AE, P-AE, SDA-AE, PL-AE) with the supervised CNN. It should be noted that for the CNN we apply the same data augmentation policy as applied for the SDA-AE and PL-AE.
\end{itemize}

\subsubsection{MNIST Results}

After training multiple Auto-Encoders with different methods and embedding sizes on the MNIST dataset, We report the best obtained accuracies in the Table \ref{results mnist} below. 

\begin{table}[H]
\resizebox{0.49\textwidth}{!}{
\centering
\begin{tabular}{ c c  c  c  c  c}
\hline
\multirow{2}{*}{\textbf{\shortstack{Embedding \\ size}}} &  \multicolumn{5}{c}{\textbf{MNIST}} \\ 
 &  \textbf{B-AE} & \textbf{P-AE} & \textbf{SDA-AE} & \textbf{PL-AE} & \textbf{CNN}\\ \hline
250 & 95.91\% & 97.78\% & 98.12\% & \textbf{98.7\%} & \textbf{99.38\%}\\ 
300 & 95.96\% & 98.37\% & 98.04\% & \textbf{99.3\%} & \textbf{99.44\%} \\ 
350 & 96.67\% & 98.49\% & 98.37\% & \textbf{99.27\%} & \textbf{99.43\%} \\ \hline
\end{tabular}
}
\caption{Best performances obtained on the test set of the MNIST dataset for different embedding sizes. The supervised CNN column is provided as reference to the other unsupervised Auto-Encoder based methods. The best unsupervised and the CNN accuracies obtained are highlighted in bold.}
\label{results mnist}
\end{table}

When looking at Table \ref{results mnist} from the effect of the embedding size perspective, it can be seen that most of the methods' accuracy reaches its peak at an embedding size of 300. Also, the PL-AE approach stores the most information in this embedding and obtains an accuracy that is almost the same as the equivalent CNN network.

From an overall accuracy perspective, It can be appreciated that the P-AE greatly improves on the accuracy of the B-AE by around 2\% whereas the contribution of the SDA-AE method compared to the P-AE is negligible. The proposed PL-AE method allows an enhancement in the performance of the unsupervised classification with 0.81\% over the P-AE \cite{p-ae} and falls short by only 0.14\% when compared to the CNN.

\begin{figure}[H]
\centering
\includegraphics[width=0.49\textwidth]{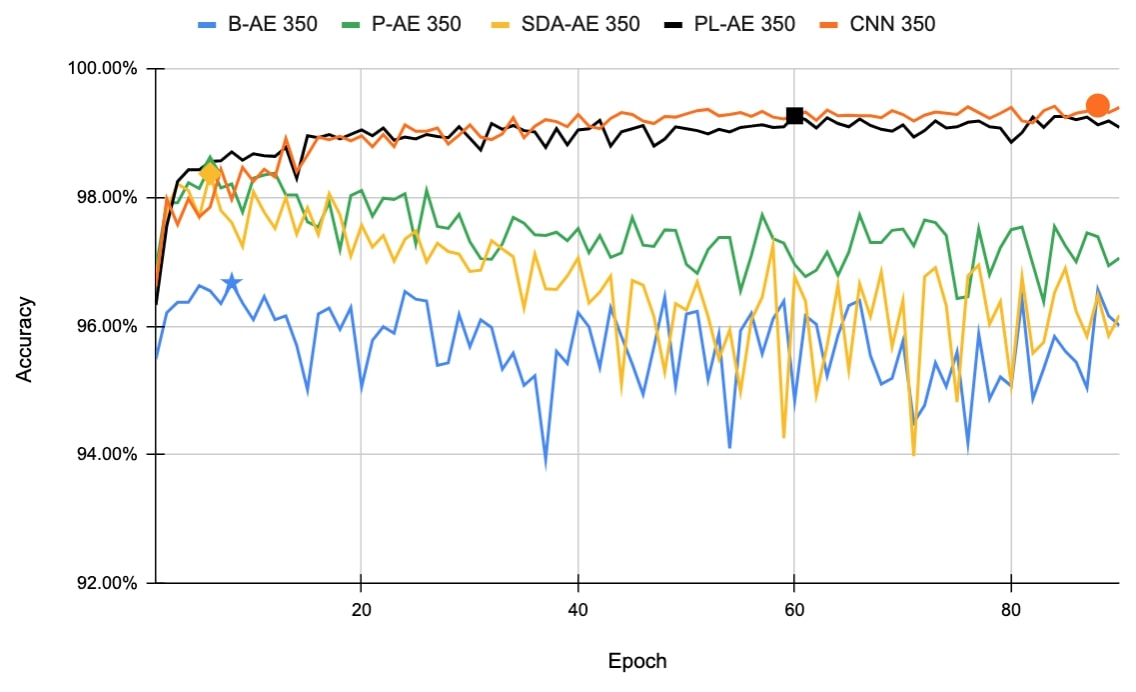}
\caption{Classification accuracy vs number of epoch on the MNIST dataset. The best accuracies obtained by the B-AE, P-AE, SDA-AE, PL-AE and the supervised CNN are marked with a blue $\bigstar$, a green $\blacktriangle$, a yellow $\blacklozenge$, a black $\blacksquare$ and an orange $\bullet$ respectively.}
\label{graph mnist}
\end{figure}

From Figure \ref{graph mnist} it can be seen that the B-AE, P-AE and the SDA-AE reach a peak in performance within the first 15 epochs then their accuracies drop. This implies that there is a loss of important information that is helpful to the classification within the generated encodings. Also, these approaches suffer from a lot of fluctuations in unsupervised classification accuracy especially after 40 epochs. As for the proposed PL-AE method, it achieves a very high and sustained performance with a maximum accuracy of 99.3\% which beats all the other Auto-Encoder based methods in both metrics i.e. accuracy and stability by a substantial margin. In addition, it can be appreciated that the PL-AE's curve is very close to the one of the CNN.

\begin{table}[H]
\resizebox{0.49\textwidth}{!}{
\centering
\begin{tabular}{ c c  c  c  c  c}
\hline
\multirow{2}{*}{\textbf{\shortstack{Line of best \\ fit parameters}}} &  \multicolumn{5}{c}{\textbf{MNIST}} \\ 
 &  \textbf{B-AE} & \textbf{P-AE} & \textbf{SDA-AE} & \textbf{PL-AE} & \textbf{CNN}\\ \hline
$m$ & -0.008 & -0.011 & -0.023 & \textbf{0.008} & \textbf{0.014} \\ 
$k$ & 96.11 & 97.95 & 97.73 & \textbf{98.54} & \textbf{98.37} \\ \hline
\end{tabular}
}
\caption{Parameters of the line of best fit of the curve classification accuracy vs number of epoch on the MNIST dataset for an embedding size of 350. The supervised CNN column is provided as reference to the other unsupervised Auto-Encoder based methods.}
\label{slope mnist}
\end{table}

The results showcased on Table \ref{slope mnist} consolidate the findings of Figure \ref{graph mnist} where it can be seen that the slope $m$ of the B-AE, P-AE and SDA-AE are null at best. Also, the $k$ parameter is lower in these methods than the PL-AE and CNN approachs. On the other hand, It can be appreciated that the PL-AE and the CNN have nearly the same parameter for the line of best fit. This proves the efficacy of the PL-AE approach w.r.t the other Auto-Encoder based methods notably the P-AE \cite{p-ae}.

\begin{figure}[H]
\centering
\includegraphics[width=0.4\textwidth]{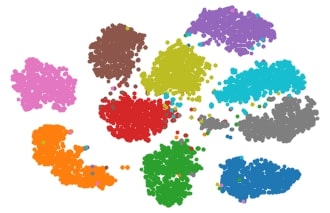}
\caption{Encodings of MNIST test images produced by the PL-AE plotted in 2D using the t-SNE algorithm. Each colour represents one class from 0 to 9.}
\label{clusters mnist}
\end{figure}

Figure \ref{clusters mnist} represents the embeddings of the test images of the MNIST dataset produced by the PL-AE and plotted in 2D using the t-SNE algorithm \cite{tsne}. It can be seen that well-formed clusters for each class are constructed which confirms the high accuracy obtained on this dataset.

\subsubsection{CIFAR-10 Results}

Table \ref{results cifar} showcases the best accuracies we obtained on the CIFAR-10 dataset for the different Auto-Encoder training methods.

\begin{table}[H]
\resizebox{0.49\textwidth}{!}{
\centering
\begin{tabular}{ c  c  c  c  c  c  }
\hline
\multirow{2}{*}{\textbf{\shortstack{Embedding \\ size}}} &  \multicolumn{5}{c}{\textbf{CIFAR-10}} \\ 
 &  \textbf{B-AE} & \textbf{P-AE} & \textbf{SDA-AE} & \textbf{PL-AE} & \textbf{CNN}\\ \hline
250 & 36.27\% & 55.22\% & 56.21\% & \textbf{60.66\%} & \textbf{80.77\%}\\ 
300 & 37.67\% & 56.33\% & 56.86\% & \textbf{60.60\%} & \textbf{81.14\%} \\ 
350 & 39.15\% & 56.46\% & 57.42\% & \textbf{60.71\%} & \textbf{81.73\%} \\ \hline
\end{tabular}
}
\caption{Best performances obtained on the test set of the CIFAR-10 dataset for different embedding sizes. The supervised CNN column is provided as reference to the other unsupervised Auto-Encoder based methods. The best unsupervised and the CNN accuracies obtained are highlighted in bold.}
\label{results cifar}
\end{table}

It can be observed from Table \ref{results cifar} that for the CIFAR-10 dataset, the PL-AE results in better accuracy compared to the P-AE and SDA-AE by 4.25\% and 3.29\% respectively which represents a substantial improvement, and achieves the best unsupervised classification accuracy with 60.71\%. Whereas when comparing the SDA-AE with the P-AE it can be seen that a maximum of almost 1\%  improvement have been made. On the other hand, the P-AE drastically improves the classification accuracy compared to the basic Auto-Encoder (B-AE) with up to 17.66\%. Finally, comparing all unsupervised methods to the CNN it is clear that the CNN still has the edge with a margin of 21.02\%.

From an embedding size perspective, it can be seen from Table \ref{results cifar} that all the methods stored the maximum information in the embeddings sizes used except for the B-AE. Thus, among all the Auto-Encoders tested the PL-AE stores the most useful information to the classification task.

\begin{figure}[H]
\centering
\includegraphics[width=0.49\textwidth]{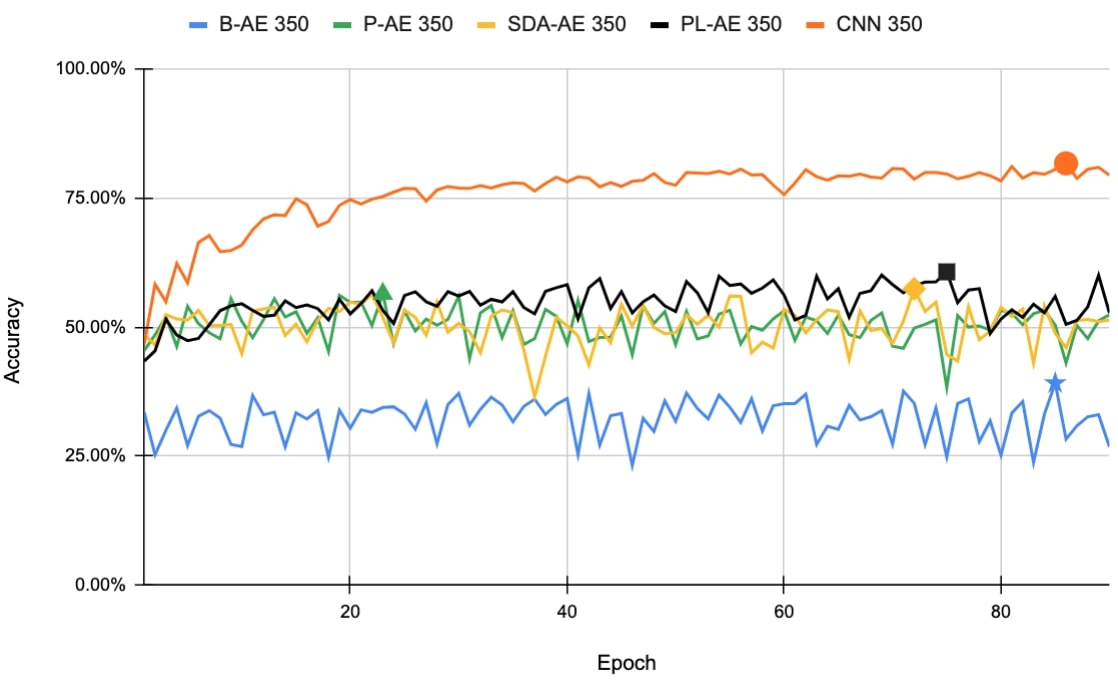}
\caption{Classification accuracy vs number of epoch on the CIFAR-10 dataset. The best accuracies obtained by the B-AE, P-AE, SDA-AE, PL-AE and the supervised CNN are marked with a blue $\bigstar$, a green $\blacktriangle$, a yellow $\blacklozenge$, a black $\blacksquare$ and an orange $\bullet$ respectively.}
\label{graph cifar}
\end{figure}

Figure \ref{graph cifar} shows that the accuracy of the B-AE, P-AE and the SDA-AE does not improve over the epochs and has a lot of fluctuations. Whereas the accuracy of the PL-AE improves over the epochs with far less fluctuations and surpasses all the other Auto-Encoder based methods.

\begin{table}[H]
\resizebox{0.49\textwidth}{!}{
\centering
\begin{tabular}{ c c  c  c  c  c}
\hline
\multirow{2}{*}{\textbf{\shortstack{Line of best \\ fit parameters}}} &  \multicolumn{5}{c}{\textbf{CIFAR-10}} \\ 
 &  \textbf{B-AE} & \textbf{P-AE} & \textbf{SDA-AE} & \textbf{PL-AE} & \textbf{CNN}\\ \hline
$m$ & 0 & -0.014 & -0.005 & \textbf{0.049} & \textbf{0.179} \\ 
$k$ & 32.24 & 51.05 & 50.63 & \textbf{52.46} & \textbf{67.71} \\ \hline
\end{tabular}
}
\caption{Parameters of the line of best fit of the curve classification accuracy vs number of epoch on the CIFAR-10 dataset for an embedding size of 350. The supervised CNN column is provided as reference to the other unsupervised Auto-Encoder based methods.}
\label{slope cifar}
\end{table}

From Table \ref{slope cifar}, it can be seen that the PL-AE approach is the only one to achieve a positive slope among the Auto-Encoder methods and it has the highest $k$ parameter whereas it still falls short against the CNN supervised approach.


\subsubsection{SVHN Results}

Table \ref{results svhn} presents the best accuracies we obtained on the SVHN dataset for the different Auto-Encoder training methods.

\begin{table}[H]
\resizebox{0.49\textwidth}{!}{
\centering
\begin{tabular}{ c  c  c  c  c  c  }
\hline
\multirow{2}{*}{\textbf{\shortstack{Embedding \\ size}}} &  \multicolumn{5}{c}{\textbf{SVHN}} \\ 
 &  \textbf{B-AE} & \textbf{P-AE} & \textbf{SDA-AE} & \textbf{PL-AE} & \textbf{CNN}\\ \hline
250 & 32.99\% & 73.01\% & 72.07\% & \textbf{75.59\%} & \textbf{92.53\%}\\ 
300 & 35.12\% & 74.67\% & 72.92\% & \textbf{76.21\%} & \textbf{92.57\%} \\ 
350 & 36.48\% & 74.76\% & 74.17\% & \textbf{76.48\%} & \textbf{92.57\%} \\ \hline 
\end{tabular}
}
\caption{Best performances obtained on the test set of the SVHN dataset for different embedding sizes. The supervised CNN column is provided as reference to the other unsupervised Auto-Encoder based methods. The best unsupervised and the CNN accuracies obtained are highlighted in bold.}
\label{results svhn}
\end{table}

It can be observed from Table \ref{results svhn} that for the SVHN dataset in terms of classification accuracy the P-AE obtains an accuracy of 74.76\% immensely improving upon the accuracy of the B-AE by a margin of 38.28\% which is more than double. The P-AE also achieves an accuracy that is higher than the one of the SDA-AE by just 0.59\%. Thus, it falls short compared to the proposed PL-AE which improves on its accuracy by almost 2\% which is also a substantial improvement given the relatively high accuracies obtained on this dataset. Still the supervised method CNN achieves the best accuracy with at least 16.36\% better accuracy.

Table \ref{results svhn} also shows that all the Auto-Encoder based methods benefited from increasing the embedding size. The P-AE, PL-AE and CNN reach their maximum accuracy at an embedding size of 300 and increasing it to 350 does not improve the accuracy much if any. Whereas the accuracy of the B-AE and the SDA-AE seem to still improve further and benefit from a larger embedding size.

\begin{table}[H]
\resizebox{0.49\textwidth}{!}{
\centering
\begin{tabular}{ c c  c  c  c  c}
\hline
\multirow{2}{*}{\textbf{\shortstack{Line of best \\ fit parameters}}} &  \multicolumn{5}{c}{\textbf{SVHN}} \\ 
 &  \textbf{B-AE} & \textbf{P-AE} & \textbf{SDA-AE} & \textbf{PL-AE} & \textbf{CNN}\\ \hline
$m$ & -0.058 & 0.049 & 0.05 & \textbf{0.106} & \textbf{0.023} \\ 
$k$ & 31.66 & 65.64 & 64.53 & \textbf{66.31} & \textbf{90.75} \\ \hline
\end{tabular}
}
\caption{Parameters of the line of best fit of the curve classification accuracy vs number of epoch on the SVHN dataset for an embedding size of 350. The supervised CNN column is provided as reference to the other unsupervised Auto-Encoder based methods.}
\label{slope svhn}
\end{table}

Table \ref{slope svhn} shows that the PL-AE approach achieves the highest parameters among all the tested methods with a slope $m = 0.106$ and a $k=66.31$ which in terms of slope is double the one of the P-AE \cite{p-ae} and SDA-AE and $5\times$ the slope of the supervised CNN. Thus, the PL-AE falls short to the supervised CNN in terms of the parameter $k$ with a 24.44 difference. 

\begin{figure}[H]
\centering
\includegraphics[width=0.49\textwidth]{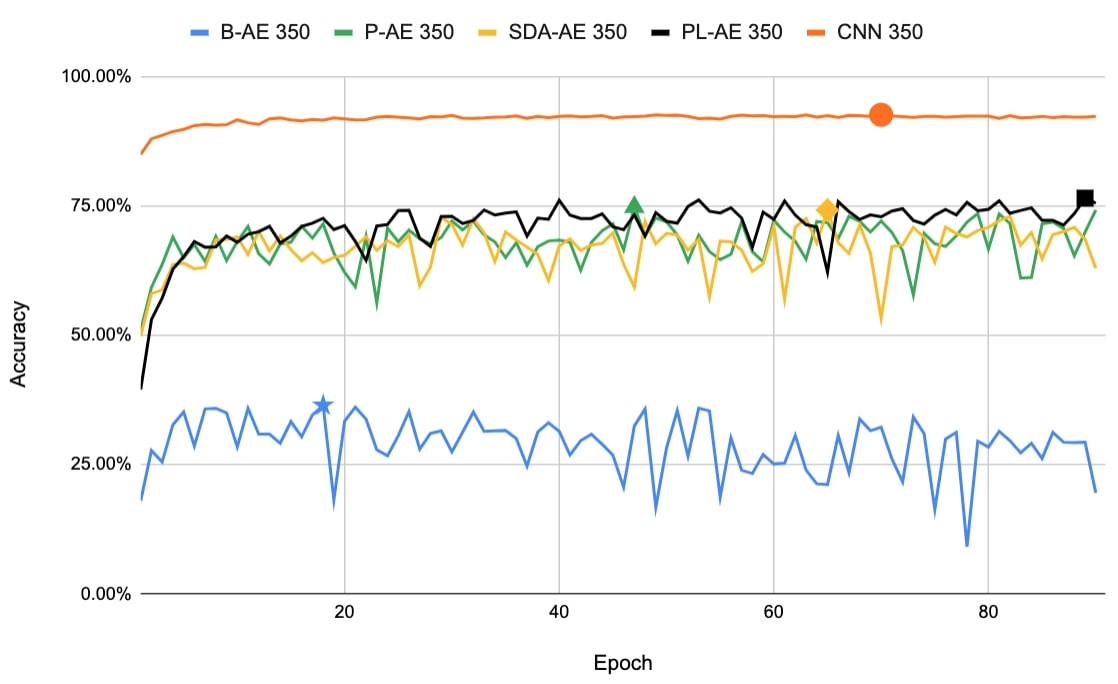}
\caption{Classification accuracy vs number of epoch on the SVHN dataset. The best accuracies obtained by the B-AE, P-AE, SDA-AE, PL-AE and the supervised CNN are marked with a blue $\bigstar$, a green $\blacktriangle$, a yellow $\blacklozenge$, a black $\blacksquare$ and an orange $\bullet$ respectively.}
\label{graph svhn}
\end{figure}

From Figure \ref{graph svhn}, it can be seen that there is no improvement in the accuracy of the P-AE and SDA-AE over the training epochs and a lot of fluctuations are present. Whereas, the PL-AE achieves a competitive performance and surpasses the other Auto-Encoder based methods after few epochs, learning better encodings with a much smoother accuracy curve. 

\section{Comparison with State of the Art}

Table \ref{final results} compares the performance of the Pseudo-Labelling method proposed in this work with other methods proposed in the literature. The performances of the other methods are reported from \cite{scae}. 

\begin{table}[H]
\resizebox{0.49\textwidth}{!}{
\centering
\begin{tabular}{  c  c  c  c  c  }
\hline
\textbf{Method} & \textbf{MNIST} & \textbf{CIFAR-10} & \textbf{SVHN} & \textbf{Reference} \\ \hline
\textbf{K-means} & 53.49\% & 20.8\% & 12.5\% & \cite{adc} \\
\textbf{AE} & 81.2\% & 31.4\% & - & \cite{ae} \\
\textbf{GAN} & 82.8\% & 31.5\% & - & \cite{unsupervised-gan} \\
\textbf{IMSAT} & 98.4\% & 45.6\% & 57.3\% & \cite{imsat} \\
\textbf{IIC} & 98.4\% & \textbf{57.6\%} & - & \cite{iic} \\
\textbf{ADC} & 98.7\% & 29.3\% & 38.6\% & \cite{adc} \\
\textbf{SCAE} & \textbf{99.0\%} & 33.48\% & \textbf{67.27\%} & \cite{scae} \\ \hline
\textbf{B-AE} & 96.67\% & 39.15\% & 36.48\% & - \\
\textbf{P-AE} & 98.49\% & 57.46\% & 74.76\% & \cite{p-ae} \\
\textbf{SDA-AE} & 98.37\% & 57.42\% & 74.17\% & Ours \\
\textbf{PL-AE} & \textbf{99.30\%} & \textbf{60.71\%} & \textbf{76.48\%} & Ours \\
\textbf{CNN} & 99.44\% & 81.73\% & 92.57\% & - \\ \hline
\end{tabular}
}
\caption{Unsupervised classification accuracy results in \%. The supervised CNN line is provided as reference to the other unsupervised methods. Previous state-of-the art accuracies and the best accuracies we obtained are highlighted in bold.}
\label{final results}
\end{table}

It can be noticed that the B-AE out-performs the K-means \cite{adc}, AE \cite{ae} and GAN \cite{unsupervised-gan} methods as it extracts encodings from data then uses it for classification as opposed to directly clustering the data. That's what allows it to have an edge over the K-means algorithm. As for the AE and the GAN they are out-performed because the B-AE uses convolutional layers vs dense layers for the AE and is more specialised than the GAN. Thus, the B-AE falls short compared to the more advanced methods like IMSAT \cite{imsat}.

The P-AE \cite{p-ae} continues where the B-AE left-off where it can be seen that it out-performs the accuracy obtained by the IMSAT and the IIC \cite{iic} on the MNIST and CIFAR-10 datasets and achieves the second best performance on the SVHN dataset. This gain in performance comes from harnessing the dependencies between the pixels in computing the loss to improve the quality of the generated embeddings. 

The SDA-AE brings no noticeable performance improvement over the P-AE even though it is based on it and incorporates data augmentation thus it seems to have a trivial impact here.

Besides, it can be appreciated that the PL-AE proposed in this work improves by 0.81\%, 3.25\% and 1.72\% over the P-AE approach \cite{p-ae} and by 0.3\%, 3.11\% and 9.21\% over the previous state-of-the-art on the MNIST, CIFAR-10 and SVHN datasets respectively. This showcases that the Pseudo-Labelling method proposed in this work incorporates data augmentation in a beneficial way which allows obtaining very high accuracies that are stable across the different datasets that have been explored. These ameliorations are accentuated by the noticeable improvements in terms of parameters of the line of best fit presented in section \ref{results}.

Also, it is important take into account that the performance and stability gains obtained by the PL-AE are using a very simple neural network architecture compared to the other approachs proposed in the literature which use very deep networks, such as in ADC \cite{adc} a ResNet-50 \cite{resnet} network is employed which is not only 50 layers deep but uses residual connections which have numerous benefits.


\section{Conclusion}
In this work, we proposed a unique combination of denoising Auto-Encoders with the perceptual loss in the pseudo-labelling method to do unsupervised image classification. The proposed method encourages the encoder to look past the corruptions that have been applied and create rich encodings that are highly informative of the input's class. We show that when data augmentation is incorporated in a simple way it does not contribute much in improving the classification accuracy. Thus, when using the Pseudo-Labelling method improvement are seen on multiple datasets in terms of accuracy, stability and the parameters of the line of best fit. State-of-the-art accuracy is obtained on the MNIST, CIFAR-10 and SVHN datasets with 99.3\%, 60.71\% and 76.48\% respectively. Also, the accuracy obtained on the MNIST dataset is lower only by 0.14\% than an equivalent supervised CNN.

A future extension to this work may include the exploration of different pre-trained models, the creation of dataset specific data augmentation transformations and the exploration of the out-of-distribution generalisation capabilities of the PL-AE.

\section*{Acknowledgement}
This work received partial funding from the European Research Council (ERC) under the European Union's Horizon 2020 research and innovation program (ERC Advanced Grant agreement No 694665 : CoBCoM - Computational Brain Connectivity Mapping) and from the French government, through the 3IA C\^{o}te d'Azur Investments in the Future project managed by the National Research Agency (ANR) with the reference number ANR-19-P3IA-0002.

\appendix
\section{Data Augmentation Transformations Parameters}
\label{params DA}
Parameters of the used data augmentation transformations :

\begin{itemize}
\item \textbf{Random rotation :} Rotating the image by up to 45 degrees clock-wise or counter clock-wise.
\item \textbf{Affine transformation :} Composed of a rotation of up to 45 degrees, scaling of the image up or down by up to 1.5$\times$ or 0.5$\times$ respectively.
\item \textbf{Crop :} Taking a random portion of the image that is of size 20$\times$20.
\item \textbf{Cutout :} Hiding one area of size 10$\times$10 from the image which corresponds to approximatively 1/9 of the image.
\item \textbf{Colour jitter} Changing the colours of the image with the following parameters brightness=0.8; contrast=0.8; saturation=0.8; hue=0.2
\end{itemize}

\section{Chosen Data Augmentation Transformations}
\label{top-10}
List of the top-10 data augmentation transformations that we chose for the CIFAR-10 and SVHN datasets.
\begin{table}[H]
\resizebox{0.49\textwidth}{!}{
\centering
\begin{tabular}{c c}
\hline
CIFAR-10 & SVHN \\ \hline
Affine transformation & Cutout \\
Cutout & Gray scale \\
Horizontal flip & Colour jitter \\
Random rotation + Affine transformation & Random rotation \\
Random rotation + Cutout & Random rotation + Gray scale \\
Random rotation + Horizontal flip & Cutout+ Gaussian noise \\
Random rotation + Vertical flip & Cutout + Gray scale \\
Affine transformation + Cutout & Cutout + Colour jitter \\
Affine transformation + Horizontal flip & Gaussian noise + Colour jitter \\
Cutout + Colour jitter & Gray scale + Colour jitter \\
\hline
\end{tabular}
}
\caption{Top-10 Chosen Data Augmentation Transformations.}
\label{}
\end{table}%

\bibliographystyle{IEEEtran}
\bibliography{IEEEabrv,refs}
  
\end{document}